\documentclass{article}
\usepackage{ijcai20}

\usepackage{times}
\usepackage{soul}
\usepackage{url}
\usepackage[hidelinks]{hyperref}
\usepackage[utf8]{inputenc}
\usepackage[small]{caption}
\usepackage{graphicx}
\usepackage{amsmath}
\usepackage{amsthm}
\usepackage{booktabs}
\usepackage{array}
\usepackage{multirow}
\usepackage{algorithm}
\usepackage{algorithmic}
\title{Overcoming Language Priors with Self-supervised Learning \\ for Visual Question Answering}

\author{
Xi Zhu$^{1,2}$ \and
Zhendong Mao$^3$\footnote{Corresponding Author}\and
Chunxiao Liu$^{1,2}$\and
Peng Zhang$^1$\and
Bin Wang$^1$\And
Yongdong Zhang$^3$
\affiliations
$^1$Institute of Information Engineering, Chinese Academy of Sciences\\
$^2$School of Cyber Security, University of Chinese Academy of Sciences\\
$^3$University of Science and Technology of China\\
\emails
\{zhuxi, liuchunxiao, pengzhang, wangbin\}@iie.ac.cn
\{zdmao,zhyd73\}@ustc.edu.cn
}

\hypersetup{draft}
\begin{document}

\maketitle

\begin{abstract}
Most Visual Question Answering (VQA) models suffer from the language prior problem, which is caused by inherent data biases. 
Specifically, VQA models tend to answer questions (\textit{e.g.}, what color is the banana?) based on the high-frequency answers (\textit{e.g.}, yellow) ignoring image contents.
Existing approaches tackle this problem by creating delicate models or introducing additional visual annotations to reduce question dependency while strengthening image dependency.
However, they are still subject to the language prior problem since the data biases have not been even alleviated.
In this paper, we introduce a self-supervised learning framework to solve this problem. 
Concretely, we first automatically generate labeled data to balance the biased data, and propose a self-supervised auxiliary task to utilize the balanced data to assist the base VQA model to overcome language priors. 
Our method can compensate for the data biases by generating balanced data without introducing external annotations.
Experimental results show that our method can significantly outperform the state-of-the-art, improving the overall accuracy from 49.50\% to 57.59\% on the most commonly used benchmark VQA-CP v2. 
In other words, we can increase the performance of annotation-based methods by 16\% without using external annotations. Our code is available in GitHub\footnote{\url{https://github.com/CrossmodalGroup/SSL-VQA}}.
\end{abstract}

\section{Introduction}
Visual Question Answering (VQA) has attracted increasing attention as an AI-complete task, whose goal is to automatically answer natural language questions according to images. The paradigm of VQA \cite{antol2015vqa,yang2016stacked,anderson2018bottom,kim2018bilinear} is to project the image and the question into a common feature space, and then fuse them as a joint vector to make prediction.
Recently, some researchers \cite{agrawal2018don,goyal2017making} have demonstrated that most existing VQA models are suffering from the language prior problem, ignoring the image contents. For example, the question ``\textit{what color is the grass?}'' can be answered by ``\textit{green}'' generally, no matter what images are given, since most corresponding answers are ``\textit{green}'' in the dataset.
As a result, the model memorizing the language priors will perform poorly on out-of-domain datasets.

Existing approaches in language prior alleviation focus on reducing question dependency while increasing image dependency,
and they can be roughly categorized as non-annotation-based methods and annotation-based methods. 
For non-annotation-based methods, researchers mostly design delicate models with different strategies. 
For example, \cite{ramakrishnan2018overcoming} proposed an adversarial learning strategy to overcome the language priors by minimizing the performance of the question-only branch adversarially. 
Rubi \cite{cadene2019rubi} reduced the influence of the most-biased instances and increased the impact of the less-biased instances by dynamically adjusting their weights. 
The annotation-based methods try to directly increase the image dependency by introducing external visual supervision.
 \cite{selvaraju2019taking} used human-attention maps to ensure the alignment between model-attention and human-attention. \cite{wu2019self} maintained the consistency of correct answers and influential objects annotated by human explanations. Typically, annotation-based methods can achieve better performance than non-annotation-based methods, since they can better understand images with the guidance of visual supervision. Nonetheless, these methods require large-scale visual annotations, which are not easily accessible.

 \begin{figure}[t]
\begin{center}
   \includegraphics[width=0.9\linewidth]{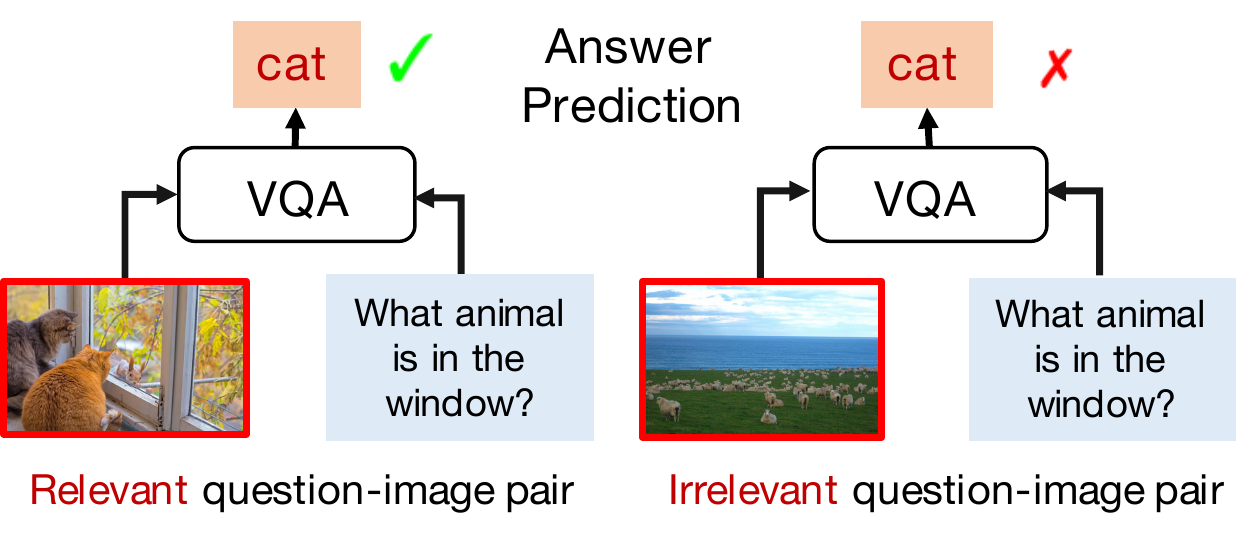}
\end{center}
   \caption{A question can only be answered based on relevant images.
   }
\label{fig:intro}
\end{figure}

However, the inherent data biases have not been even alleviated, and the above methods just weaken the adverse effect of them to some extent and hence yield unsatisfactory performance. The inherent data biases will inevitably force the VQA model to be inclined to the high-frequency answers with higher confidence and eventually arouse language prior problem.
Therefore, it is of crucial importance to solve the inherent data biases, i.e., transforming biased data to balanced data without introducing external annotations.

To this end, we propose a self-supervised learning framework for VQA to automatically balance the biased data to overcome language prior problem.
Our method is motivated by an interesting and intuitive finding. As shown in Figure \ref{fig:intro}, a question can only be answered when the given image containing the key information for answering the question. We define such a question-image pair as relevant, it is irrelevant otherwise.
Based on the above observation, it is necessary to estimate whether the given question and image are relevant or not before answering the question.
For that purpose, we introduce an auxiliary task named \textit{question-image correlation estimation} to estimate the relevance between questions and images.
Specifically, we first automatically generate a set of balanced question-image pairs with binary labels (relevant and irrelevant), which are then consumed by the self-supervised auxiliary task to assist the VQA model to overcome language priors.
We incorporate the auxiliary task into the base VQA model by feeding the relevant and irrelevant pairs.
When fed a relevant question-image pair, the VQA model is encouraged to predict the correct answer with a high confidence score, where the confidence score is the probability of the question-image pair being relevant.
On the contrary, the VQA model is pushed to predict the correct answer with a low confidence score when the input pair is irrelevant.
Moreover, the confidence scores of irrelevant pairs can be used as a guide to measuring the language priors, which can avoid over-fitting. 
By optimizing these two objectives simultaneously, we can achieve a balance between answering questions and overcoming language priors.
Therefore, our method can also be interpreted as an underlying multi-task learning framework.

To summarize, our contributions are as follows:
\begin{itemize}
\item We introduce a self-supervised framework by transforming the inherently biased data into balanced data automatically, and propose an auxiliary task to exploit such balanced data to overcome language priors fundamentally. To the best of our knowledge, this is the first work to use self-supervised learning in this domain.
\item Extensive experiments are conducted on the popular benchmark VQA-CP v2. Experimental results show that our approach without using external annotations can significantly outperform the state-of-the-art, including the models using human supervision. We increase the overall accuracy from 49.50\% to 57.59\%.
\end{itemize}

\section{Related Works}
\subsection{Visual Question Answering}
Visual Question Answering (VQA) aims to answer questions according to images, 
which involves technologies from both natural language processing and computer vision communities \cite{liu2016hierarchical,parkhi2015deep,conneau2016very,ijcai2018-114}. 
Existing VQA approaches can be coarsely classified into four categories: 
1) Joint Embedding approaches \cite{antol2015vqa} first project images and questions into a common feature space, and then combine them to predict answers by a classifier. 
2) Attention-based methods \cite{anderson2018bottom} mainly focus on learning the interactions between the question words and image regions, making the answering process to be more interpretable. 
3) Compositional models \cite{andreas2016neural} leverage the compositional structure of questions to assembling modules that operate in the space of attention. 
4) Knowledge-based approaches \cite{wu2016ask} are proposed to answer common sense questions by exploiting external knowledge.

However, existing models tend to memorize the language priors during training without considering image information. 
Such models may achieve unexpectedly impressive results on the test set sharing the same distribution with the training set, but often performs poorly on out-of-domain test sets.

\begin{figure*}[!t]
\begin{center}
\includegraphics[width=1\linewidth]{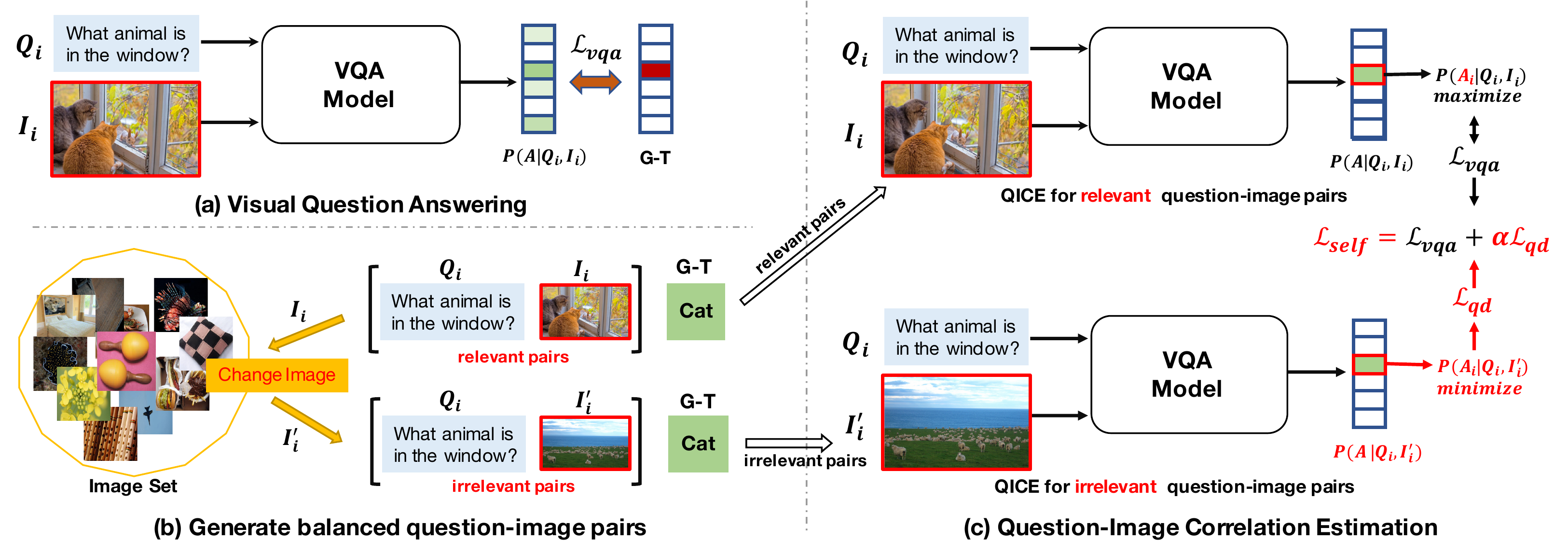}
\end{center}
   \caption{The framework of our self-supervised approach. The base VQA model is depicted in part (a), which aims to answer a question according to an image. (b) displays how we automatically generate balanced question-image pairs. To be more clear, (c) shows how the question-image correlation estimation works for relevant and irrelevant pairs separately. G-T denotes the ground truth.
   }
\label{fig:model}
\end{figure*}

\subsection{Overcoming Language Priors in VQA}

Existing approaches aiming at overcoming language priors can be roughly categorized as non-annotation-based methods and annotation-based methods. 
Non-annotation-based methods focus on creating delicate models to reduce the question dependency, while the annotation-based methods concentrate on strengthening the visual grounding by introducing additional human visual supervision.

For the non-annotation-based methods, \cite{agrawal2018don} proposed a hand-designed VQA framework, which explicitly disentangled the visual recognition from answer space prediction for different question types. Similarly, \cite{jing2020overcoming} also decoupled the concept discovery and the question answering. 
Apart from shrinking the answer space, \cite{ramakrishnan2018overcoming} proposed an adversarial learning strategy by minimizing the performance of the question-only branch adversarially. \cite{guo2019quantifying} adopted a pair-wise ranking schema, forcing the question-only branch to make worse predictions than the base model did.
Rubi \cite{cadene2019rubi} dynamically adjusted the weights of training instances via its prior masks learned by the question-only branch, reducing the influence of the most-biased instances and increase the impact of the less-biased instances. 
\cite{yi2018neural} proposed a neural-symbolic model incorporating the symbolic program executor into DNN for visual reasoning, which is distinct from the above models and can also solve the bias problem. 
\cite{mao2019neuro} combined the neural-symbolic model with curriculum concept learning, making it more generalizable.

Beyond that, annotation-based methods are shown to be effective by highlighting the important visual regions under the guidance of external visual supervision. HINT \cite{selvaraju2019taking} increased the image dependency by optimizing the alignment between human-attention maps and gradient-based visual importance. SCR \cite{wu2019self} also emphasized the correspondences between correct answers and influential objects annotated by human textual explanations. However, these models are heavily dependent on human supervision, which is not always accessible.

Different from all these methods, our self-supervised approach does not need to construct complex architectures or introduce external supervision. 
We first balance the original biased data by automatically generating balanced labels, and then overcome the language priors based on the balanced data with an auxiliary task in a self-supervised manner.

\subsection{Self-supervised Learning}
Self-supervised learning constructs some supervisory signals automatically computed from the input data, and efficiently exploits the input itself to learn the high-level representation for some down-stream tasks.  For example, \cite {gidaris2018unsupervised} proposed to randomly rotate an image by one of four possible angles and let the model predict that rotation. Apart from trying to predict the rotation, one can also try to recover part of the data, such as image completion \cite{pathak2016context}.
In this paper, we utilize self-supervised learning for question-image correlation estimation as an auxiliary task to assist VQA models to overcome language priors. We randomly change the image in the original relevant question-image pair, and then let the model predict its relevance. 

\section{Method}
The framework of our approach is illustrated in Figure \ref{fig:model}. Next, we will make detailed description of how it works.

\subsection{The Paradigm of VQA}
\label{Preliminaries}
The purpose of VQA is to automatically answer textual questions according to images. 
Concretely, given a VQA dataset $\mathcal{D} = \{\mathit{I}_i, \mathit{Q}_i, \mathit{A_i}\}_{i=1}^N$ with N instances, where $\mathit{I}_i \in \mathcal{I}$, $\mathit{Q}_i \in \mathcal{Q}$ are the image and question for the $i^{th}$ instance while $A_i \in \mathcal{A}$ is annotation, the VQA model aims to learn a mapping function $\mathcal{F} : \mathcal{I} \times \mathcal{Q} \to \mathcal{R}^{\mathcal{A}}$ to produce an accurate distribution over the answer space $\mathcal{A}$. 
It typically consists of three parts: extracting features for both image and question, fusing them to obtain a joint multi-modal representation, and predicting a distribution over the answer space. Consequently, we can write the answer prediction for the $i^{th}$ image and question as $\mathcal{F}(\mathcal{A}|\mathit{I}_i, \mathit{Q}_i)$.
Almost all the existing VQA models \cite{yang2016stacked,kim2018bilinear,anderson2018bottom} follow this paradigm and their parameters are typically optimized by minimizing the cross-entropy loss $\mathcal{L}_{vqa\_ce}$ in Equation~(\ref{equ:vqa_ce}) or multi-label soft loss $\mathcal{L}_{vqa\_ml}$ in Equation~(\ref{equ:vqa_loss_soft}).

\begin{equation}
\mathit{P}(\mathcal{A}|\mathit{I}_i, \mathit{Q}_i) = softmax(\mathcal{F}(\mathit{I}_i, \mathit{Q}_i)) 
\end{equation}

\begin{equation}
\mathcal{L}_{vqa\_ce} = -\frac{1}{N} \sum _i^N{log\mathit{P}(\mathcal{A}|\mathit{I}_i, \mathit{Q}_i)[\mathit{A_i}]}
\label{equ:vqa_ce}
\end{equation}

\begin{equation}
\begin{aligned}
\mathcal{L}_{vqa\_ml} &= -\frac{1}{N} \sum _i^N{[\mathit{t_i}log(\delta(\mathcal{F}(\mathcal{A}|\mathit{I}_i, \mathit{Q}_i)))} \\ 
& + (\mathit{1-t_i})log(1-\delta(\mathcal{F}(\mathcal{A}|\mathit{I}_i, \mathit{Q}_i))] 
\label{equ:vqa_loss_soft}
\end{aligned}
\end{equation}
where $\delta(\cdot)$ denotes sigmoid function. And $t_i$ is the soft target score of each answer for the $i^{th}$ instance, denoted as $t_i = {{number~ of ~votes}\over{n}}$, where $n$ is the number of valid answers for the $i^{th}$ question, and ${number~ of ~votes}$ is the number of each answer that human annotated for that question. 

\subsection{Question-Image Correlation Estimation}
A VQA model memorizing the language priors tends to make predictions directly ignoring the image.
Ideally, a question can only be answered when the given image containing the information related to it.
Therefore, it is of crucial importance to require the VQA model to judge whether the given image can be used as the reference or not before answering a specific question, which has been neglected by almost all the previous works since all the question-image pairs are matched correctly in existing benchmarks. 
We illustrate that such validation is necessary to alleviate language priors in VQA, because it can force the model to refer image contents rather than answer blindly. 
To this end, we propose an auxiliary task called Question-Image Correlation Estimation (QICE), a binary classification task, to predict whether a question-image pair is relevant or not before answering the question. 
In this paper, we define the relevant question-image pair as the image can be used to answer the question with a specific answer. 

\paragraph{Generate balanced question-image pairs.} 
We first automatically generate a set of labeled question-image pairs from the original dataset without human annotations for the auxiliary task as shown in Figure \ref{fig:model} (b). Specifically, each question-image pair $(Q,I)$ in the training set is treated as a relevant pair with label $c=1$, because there is an answer $A$ for this pair in the dataset. And then for each relevant pair $(Q,I)$, we replace the original image $I$ by a randomly selected image from the image set $\mathcal{I}$, which is denoted as $I^{\prime}=Sample(\mathcal{I}\backslash \text{I})$.
In this way, we can get another question-image pair $(Q,I^{\prime})$. 
Obviously, the probability of $(Q,I^{\prime})$ being a relevant pair is very low considering the huge size of $\mathcal{I}$, thus we assign an irrelevant label $c=0$ to each generated pair. 
As a result, we can obtain a balanced question-image pair matching dataset where the number of relevant pairs is equal to that of the irrelevant pairs.
Note that the construction of balanced data does not need any human annotation.

\paragraph{Correlation estimation.}
With the generated balanced data, we can train the QICE model to predict the relevant label of each question-image pair by optimizing cross-entropy loss.
\begin{equation}
\begin{aligned}
\mathcal{L}_{self} &= -\frac{1}{2N}\sum_{i}^{2\mathit{N}} c_i log\mathit{QICE}(Q_i,I_i) \\
&+ (1-c_i)log(1-\mathit{QICE}(Q_i,I_i) ]
\end{aligned}
\end{equation}
$\mathcal{L}_{self}$ can be interpreted as a self-supervised training loss since it only leverages the label supervision $c$ from our generated data.
The objective function guarantees the QICE model to understand the question as well as image contents because each $Q$ corresponds to balanced relevant and irrelevant instances and no language priors can be depended on.
In the next subsection, we will discuss how to leverage our auxiliary task QICE with the balanced data to assist the VQA model to eliminate language biases in a unified framework.

\subsection{Unified Self-supervised Framework}
In this section, we present a unified VQA framework that can answer questions and estimate question-image relevance simultaneously during training. 
Obviously, the QICE task defined above can share the same network structure with VQA because they have the completely same inputs and similar outputs: they all take question-image pair $(I,Q)$ as input, and VQA predicts a distribution over answer space $\mathcal{A}$ while QICE produces a binary label on a specific answer $A$. 
Such property motivates us to settle these two tasks concurrently in a unified VQA framework as shown in Figure \ref{fig:model}. 

For the VQA model depicted in Figure \ref{fig:model} (a), it takes a relevant question-image pair $(Q,I)$ as input, and  predict a distribution $\mathcal{F}(\mathcal{A}|Q,I)$ over answer space $\mathcal{A}$, which can be optimized by minimizing VQA loss  $\mathcal{L}_{vqa\_ce}$ or $\mathcal{L}_{vqa\_ml}$. 
This objective function teaches the model to learn the capability of answering questions. 
For QICE displayed in Figure \ref{fig:model} (c), given a question-image pair $(I,Q)$ corresponding to a specific answer $A$, the prediction probability $P(A|Q,I)$ of the VQA model can be regarded as the confidence of $(I,Q)$ being a relevant pair. The larger the probability, the higher the matching degree. Therefore, $\mathcal{L}_{self}$ can be rewritten as: 
\begin{equation}
\begin{aligned}
\mathcal{L}_{self} &= -\frac{1}{2N}\sum_{i}^{2\mathit{N}} [c_i log\mathit{P}(A_i|Q_i,I_i) \\
&+ (1-c_i)log(1-\mathit{P}(A_i|Q_i,I_i)) ]
\end{aligned}
\end{equation}

The model is required to make the right binary predictions for question-image correlation estimation task, which can enforce the model to better understand images since each question is paired with equal amounts of relevant and irrelevant images.
More specifically, the first term of $\mathcal{L}_{self}$ aims to maximize the confidence of a question-image pair to be relevant,
which is consistent with the objective of the VQA task that makes a prediction on ground-truth $A$ with high confidence.

What's most important, the second term of $\mathcal{L}_{self}$ is designed to minimize the confidence of a pair to be relevant, which can exactly meet with the language priors reduction.
Intuitively, the question dependency of a VQA model can be measured by the confidence of the question been answered correctly when given irrelevant images. 
The larger the confidence, the stronger the dependency. Minimizing the confidence of irrelevant pairs being relevant can explicitly prevent the VQA model from being overly driven by the language priors, and here we name it as question dependency loss $\mathcal{L}_{qd}$:
\begin{equation}
\mathcal{L}_{qd} = -\frac{1}{N}\sum_{i}^\mathit{N}log(1-\mathit{P}(A_i|Q_i,I_i^{\prime}))
\label{equ:l_qd_1}
\end{equation}
We omit $c_i$ in Equation (\ref{equ:l_qd_1}) since $\mathcal{L}_{qd}$ is only valid for irrelevant question-image pairs $(Q,I^{\prime})$. 
Mathematically, minimizing $-log(1-\mathit{P}(A|Q,I^{\prime}))$ in $\mathcal{L}_{qd}$ is equivalent to minimizing $\mathit{P}(A|Q,I^{\prime})$. 
Experimentally, minimizing $\mathit{P}(A|Q,I^{\prime})$ is more stable than minimizing $-log(1-\mathit{P}(A|Q,I^{\prime}))$ during training, which is because the gradient of $P(A | Q,I^{\prime})$ is more stable than that of $-log(1-\mathit{P}(A|Q,I^{\prime}))$.
Therefore, in this paper, we directly optimize $\mathit{P}(A|Q,I^{\prime})$ instead, and the question dependency loss $\mathcal{L}_{qd}$ can be updated as below: 
\begin{equation}
\mathcal{L}_{qd} = \frac{1}{N}\sum_{i}^\mathit{N}\mathit{P}(A_i|Q_i,I_i^{\prime})
\end{equation}

As a result, the QICE task can be naturally regarded as underlying multi-task learning, containing two tasks: the original VQA task and the language priors reduction task. We can reformulate $ \mathcal{L}_{self}$ as following: 
\begin{equation}
\mathcal{L}_{self}=\mathcal{L}_{vqa}+\alpha\mathcal{L}_{qd}
\end{equation}
where $\mathcal{L}_{vqa}$ can be any VQA loss ($\mathcal{L}_{vqa\_ce}$ or $\mathcal{L}_{vqa\_ml}$), and $\alpha$ is a hyper-parameter.
Obviously, $\mathcal{L}_{self}$ can be seen as a generalized VQA loss, as it degenerates to $\mathcal{L}_{vqa}$ when $\alpha=0$.
That means the question dependency loss $\mathcal{L}_{qd}$ actually acts as a regularizer, preventing the VQA model from memorizing the language priors and forcing it to better understand images.
As a result, $\mathcal{L}_{self}$ offers flexibility in controlling the balance between answering questions and reducing language priors.
Moreover, we do not need to explicitly optimize the model to be expert in estimating the correlations of question-image pairs, and we just use its balanced supervision to compensate for the biases in VQA with our self-supervised loss. 
Following this, our method can alleviate language priors in a self-supervised manner without using external supervision.  

\section{Experiments}
\subsection{Datasets and Baselines}
\paragraph{Datasets.}  Our approach is evaluated on the most commonly used benchmark VQA-CP v2 \cite{agrawal2018don} using the standard evaluation metric \cite{antol2015vqa}. The VQA-CP v2 dataset is derived from VQA v2 \cite{goyal2017making} by re-organizing the train and validation splits, and the QA pairs in the training set and test set have different distributions.
Therefore, it is suitable for evaluating the model's generalizability.
We also evaluate our model on the VQA v2 dataset containing strong biases and report the results on its validation split.

\paragraph{Baselines.}
Our approach is compared with several methods, including (1) non-annotation-based methods: UpDn \cite{anderson2018bottom}, AdvReg \cite{ramakrishnan2018overcoming}, Rubi \cite{anderson2018bottom} and DLR \cite{jing2020overcoming}; (2) annotation-based methods: HINT \cite{selvaraju2019taking} and SCR (best-performing method) \cite{wu2019self}.

\subsection{Implementations Details}
Our approach is model agnostic and can be well applied to different VQA models. In this paper, we mainly evaluate our method based on UpDn \cite{anderson2018bottom}, and we add one Batch Normalization layer before the classifier. Following baselines, we use the pre-trained Faster R-CNN to extract image features. For each image, it is encoded as a set of 36 objects with corresponding 2048-dimensional feature vector. All the questions are padded to the same length 14. For each question, the words are initialized by the 300-dimensional Glove embeddings and then feed into GRU to get a sentence-level representation with the dimension of 1280.

We pre-train the model with the VQA loss for 12 epochs and fine-tune it with the self-supervised loss for 20 epochs. The batch size is 256, and the irrelevant images are randomly selected from mini-batches. The Adam optimizer is adopted with the initial learning rate of 0.001 which is halved every 5 epochs after 10 epochs.
We evaluate our approach with different VQA losses in our main experiment, setting $\alpha=3$ for multi-label VQA loss and $\alpha=1.2$ for cross-entropy VQA loss. All the other experiments in this paper are based on multi-label VQA loss with $\alpha=3$. The hyper-parameter $\alpha$ setting is also investigated in the next subsection.

\begin{table}[!t]
\centering
\begin{tabular}{lccccc}
\toprule
Method      & Yes/No & Num   & Other & Overall \\
\midrule
UpDn \shortcite{anderson2018bottom}                            & 42.27  & 11.93 & 46.05 & 39.74 \\ 
AdvReg \shortcite{ramakrishnan2018overcoming}                   & 65.49  & 15.48 & 35.48 & 41.17   \\
Rubi \shortcite{cadene2019rubi}                         & 68.65      & 20.28     & 43.18   & 47.11\\
DLR  \shortcite{jing2020overcoming} & 70.99 &18.72	 &45.57 &48.87\\
\midrule 
HINT \shortcite{selvaraju2019taking}                         & 70.04      & 10.68     & 46.31 & 47.70\\
SCR \shortcite{wu2019self}                         & 71.60      & 11.30     & 48.40     & 49.50\\
\midrule
UpDn$^\dagger$-${\mathcal{L}_{ce}}$    & 47.27  & 13.67  &  40.32  & 38.28\\
UpDn$^\dagger$-${\mathcal{L}_{ml}}$  &      43.45  & 13.64  &  48.18  & 41.53\\
\textbf{UpDn+Ours}-${\mathcal{L}_{{ce}}}$       & \textbf{87.75}     &  26.40     & 41.42    & 52.63\\
\textbf{UpDn+Ours}-${\mathcal{L}_{ml}}$       & 86.53     &  \textbf{29.87}     & \textbf{50.03}    & \textbf{57.59}\\
\bottomrule
\end{tabular}
\centering
\caption{Performance on VQA-CP v2 test split. The first row shows the performance of non-annotation-based models, while the second row displays that of annotation-based methods. Our method significantly outperforms all these methods (including the  best-performing method) no matter which VQA loss is used. $^\dagger$ denotes the reimplementation of our baseline. $\mathcal{L}_{ce}$ is cross-entropy VQA loss and $\mathcal{L}_{ml}$ is multi-label VQA loss. Accuracies in percentage (\%) are reported. }
\label{tab:1} 
\end{table}

\subsection{Experimental Results and Analysis}
\label{experiments} 

\paragraph{Comparison with state-of-the-art.}  
Our approach is tested based on two VQA losses (cross-entropy loss and multi-label loss) separately. To eliminate the stochasticity from the random sampling strategy, we report an average score of 10 experiments on the test set. 
From the results shown in Table \ref{tab:1}, we can observe that: 
(1) Our approach can not only improve the overall performance of the baseline UpDn (\textbf{+14.35\%} for cross-entropy loss and \textbf{+16.06\%} for multi-label loss), but also significantly outperform the best-performing method SCR 
(\textbf{+3.13\%} for cross-entropy loss and \textbf{+8.09\%} for multi-label loss).
(2) The improvements based on both VQA losses are all remarkable. Typically, using multi-label loss can achieve better performance since it is consistent with the evaluation metric and considers multiple feasible answers, which is shown to be more generalizable. 
(3) No matter which VQA loss is used, we can achieve extremely high accuracy (\textbf{87.75\%} and \textbf{86.53\%}) on the ``Yes/No'' question type, which indicates that our strategy is indeed to be effective in overcoming the language priors since the biases are more likely to exist in these simple questions. 
(4) For the hardest ``Num'' questions, we can also get surprising improvements, which strongly illustrates that our approach can jointly understand images and questions, and reason them efficiently. 

\begin{table}[!t]
\begin{center}
\begin{tabular}{lccccc}
\toprule
\multirow{2}{*}{Model} & \multicolumn{5}{c}{Proportion of Training Set}  \\\cline{2-6}  
                       & 20\%       & 40\%    & 60\%      & 80\%      & 100\%     \\ \midrule 
UpDn$^\dagger$\shortcite{anderson2018bottom}                   & 36.22         & 38.90      & 39.40        & 40.61       & 41.53       \\ 

SCR \shortcite{wu2019self}                  & -         & -      & -        & -       & 49.50       \\ 
\midrule
\textbf{UpDn+Ours}                   & \textbf{52.71}    & \textbf{54.42}     & \textbf{56.83}   & \textbf{57.31}   & \textbf{57.59}   \\ 
\bottomrule
\end{tabular}
\end{center}
\caption{Performance on VQA v2 test set with various amounts of training data. Our approach outperforms the baseline UpDn with an average improvement of +16.6\%. $^\dagger$ is the reimplementation based on the released code. Overall accuracies in percentage (\%) are reported.} 
\label{tab:2}
\end{table}

\paragraph{Performance on smaller training sets.}
To further demonstrate the advantage of our approach, we conduct a series of experiments based on different amounts of training data sampling from the original training set randomly. 
All the experiments are tested on the standard test set and results are shown in Table~\ref{tab:2}. We find that our method gets an average accuracy improvement of \textbf{+16.6\%} over baseline UpDn.
What's most important, even with 20\% of the training data, our approach can also significantly surpass the best-performing method SCR trained with external supervision on the full training set. We believe that is because our approach can effectively leverage the balanced data with the assistance of our regularizer, which is more likely to exhibit great generalizability.

\paragraph{Performance based on different baselines.} 
We also conduct experiments based on two additional VQA models: SAN \cite{yang2016stacked} and BAN \cite{kim2018bilinear}. 
From the results depicted in Table~\ref{tab:3}, we can observe that the improvements for different baselines are all remarkable and consistent, which demonstrates that our method is model agnostic. 

\begin{figure}[t]
\begin{center}
\includegraphics[width=0.95\linewidth]{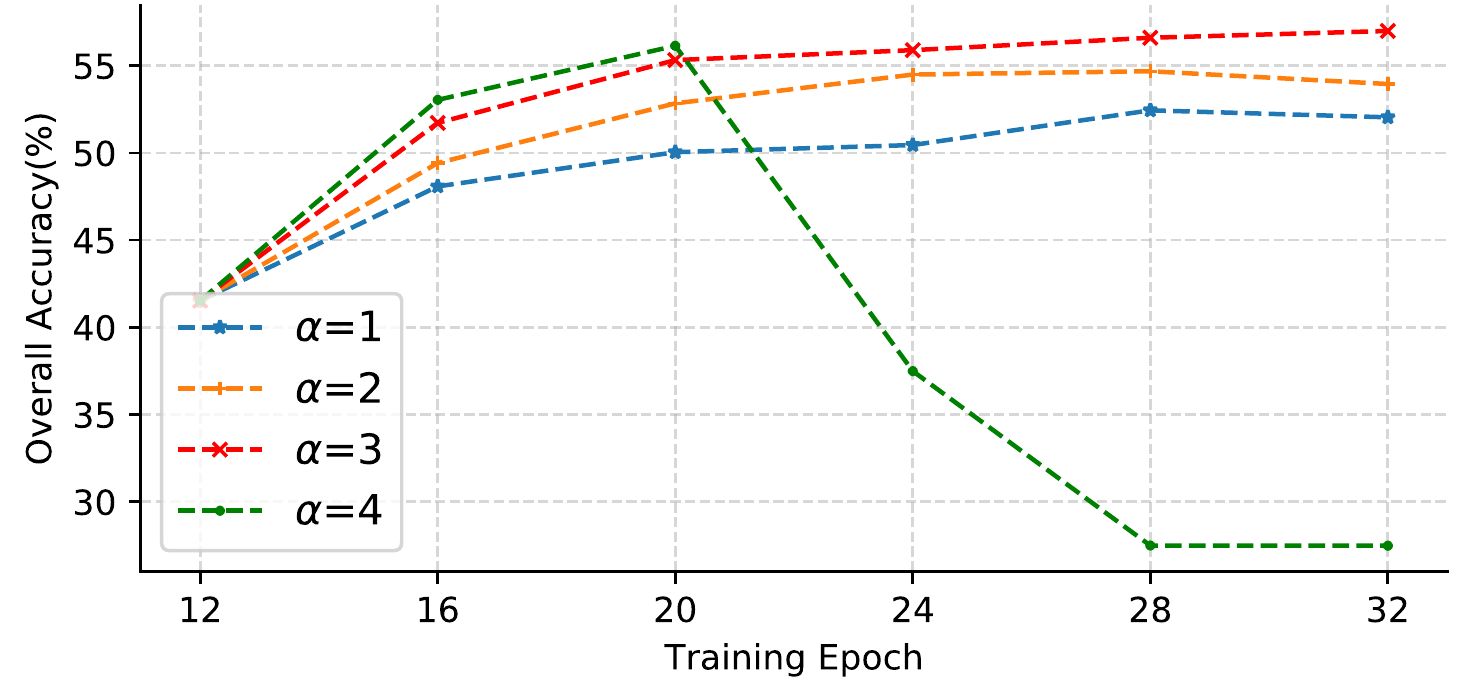}
\end{center}
   \caption{Comparison of overall accuracies with different $\alpha$ settings. Our method achieves better performance when $\alpha=3$.}
\label{fig:alpha}
\end{figure}

\begin{table}[t]
\centering
\begin{tabular}{lcc}
\toprule
\multirow{1}{*}{Method} & Overall &  Gap$\Delta\uparrow$  \\ \midrule
SAN \cite{yang2016stacked}                      & 24.96 & \multirow{2}{*}{\textbf{+12.68}} \\ 
\textbf{SAN+Ours}                & \textbf{37.64}    \\
\midrule
BAN \cite{selvaraju2019taking}              & 41.48 &\multirow{2}{*}{\textbf{+13.48}}   \\
\textbf{BAN+Ours}             & \textbf{54.96}  \\
\bottomrule
\end{tabular}
\centering
\caption{Performance on VQA v2 test set based on different baselines. Overall accuracies in percentage (\%) are reported.}
\label{tab:3} 
\end{table}

\begin{figure}[t]
\begin{center}
\includegraphics[width=0.9\linewidth]{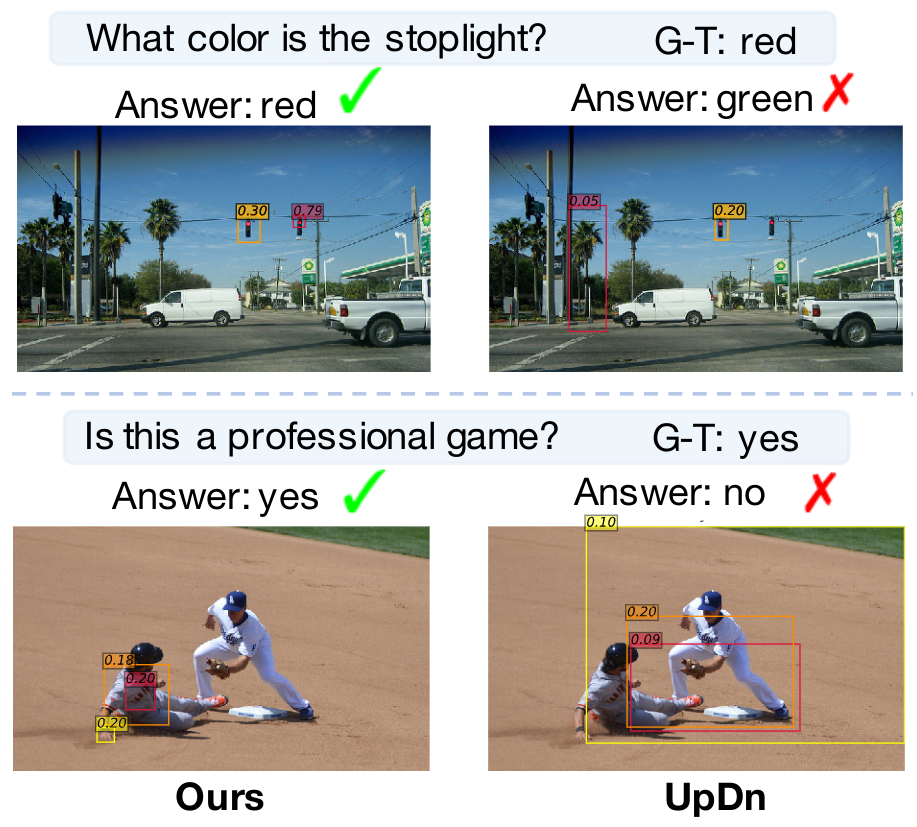}
\end{center}
   \caption{Qualitative comparisons between our self-supervised approach and the baseline UpDn. The bounding boxes indicate the most important regions with attention values. G-T is ground-truth.}
\label{fig:case}
\end{figure}

\begin{table}[t]
\centering
\begin{tabular}{lc}
\toprule
\multirow{1}{*}{Method} & Overall   \\ \midrule
UpDn \cite{anderson2018bottom}                      & 63.48 \\ 
AdvReg \cite{ramakrishnan2018overcoming}                & 62.75    \\
DLR  \cite{jing2020overcoming}   &57.96 \\
HINT \cite{selvaraju2019taking}              & 62.35   \\
SCR \cite{wu2019self}                    		  & 62.20 \\
\midrule
\textbf{UpDn+Ours}             & \textbf{63.73}  \\
\bottomrule
\end{tabular}
\centering
\caption{Overall accuracy(\%) on VQA v2 val set. Our approach does not hurt the performance of the model on the biased dataset, while other bias-reducing methods all get performance drops.}
\label{tab:4} 
\end{table}

\paragraph{Performance on biased VQA dataset.}
We also evaluate our approach on VQA v2 dataset containing strong language biases.
We pre-train the model with VQA loss for 6 epochs and then fine-tune it for 10 epochs. 
The result is shown in Table \ref{tab:4}. Our approach gets an accuracy improvement on VQA v2 val, while others may result in drops. The reason behind this is that our self-supervised loss can achieve a balance between answering questions and eliminating language priors.

\paragraph{Impact of different $\alpha$.} 
To investigate the impact of the hyper-parameter $\alpha$, which makes a trade-off between answering questions and overcoming language priors, we conduct extensive experiments with different $\alpha$ settings. 
Due to space limitations, in this paper, we only analyze the case when using multi-label VQA loss, see Figure~\ref{fig:alpha}. The model yields higher performance when $\alpha=3$. What's more, a large $\alpha$ might cause model collapse after several epochs, while a small $\alpha$ will result in unsatisfactory performance. 

\paragraph{Qualitative analysis.} 
We quantitatively evaluate the effectiveness of our approach. As shown in Figure~\ref{fig:case}, our method can answer the questions correctly and focus on the right regions. For example, when answering the question ``Is this a professional game?'', our method can pay more attention to the characters on the man's clothes, which might be an important visual clue to judge whether the game is professional.

\section{Conclusion} 
In this paper, we propose a novel self-supervised learning framework to overcome language priors in VQA. 
Based on a model-agnostic auxiliary task, our framework is able to effectively exploit the automatically generated balanced data to alleviate the influence of dataset biases.  
Experimental results show that our approach achieves a balance between answering questions and overcoming language priors, and leads to a better overall learning outcome, achieving a new state-of-the-art on the most commonly used benchmark VQA-CP v2. 
Theoretically, we believe that our work can be a meaningful step in realistic VQA and solving the language bias issue, and this self-supervision can be generalized to other tasks (e.g. image caption) that are subject to the inherent data biases.

\section*{Acknowledgments}
We thank all the anonymous reviewers for their valuable comments.
This work is supported by the National Natural Science Foundation of China (grant No.U19A2057), the National Science Fund for Distinguished Young Scholars (grant No.61525206), the Fundamental Research Funds for the Central Universities (grant No.WK3480000008), the Strategic Priority Research Program of Chinese Academy of
Sciences (grant No.XDC02040400), the National Key Research and Development Program  (grant No.2016QY03D0503) and Tianjin New Generation Artificial Intelligence Major Program (grant No.19ZXZNGX00110).

\bibliographystyle{named}
\bibliography{ijcai20}

\end{document}